\documentclass[10pt,journal,compsoc]{IEEEtran}

\IEEEoverridecommandlockouts

\usepackage{url}
\usepackage{amssymb}
\usepackage{graphicx}
\usepackage{algorithm,algorithmic}
\usepackage{tikz}
\usepackage{mathtools}
\usepackage[compress]{cite}
\usepackage{amsmath}
\usepackage{dsfont}
\usepackage{xfrac}
\usepackage{stmaryrd}
\usepackage{amsmath}
\usepackage{amssymb}
\usepackage{psfrag}

\PassOptionsToPackage{usenames,dvipsnames,svgnames}{xcolor}
\usetikzlibrary{arrows,positioning,automata,shapes,calc,shadows}

\usepackage{tcolorbox}
\definecolor{mycolor}{rgb}{0.122, 0.435, 0.698}
\makeatletter
\newcommand{\mybox}[1]{%
  \setbox0=\hbox{#1}%
  \setlength{\@tempdima}{\dimexpr\wd0+13pt}%
  \begin{tcolorbox}[colframe=black,boxrule=0.5pt,arc=4pt,left=6pt,right=6pt,top=6pt,bottom=6pt,boxsep=0pt,width=\@tempdima]
    #1
  \end{tcolorbox}
}
\makeatother

\newtheorem{proposition}{Proposition}
\newtheorem{definition}{Definition}

\newtheorem{remark}{Remark}

\newtheorem{assumption}{Assumption}
\newtheorem{theorem}{Theorem}

\newcommand{\trace}{\text{trace}}

\definecolor{matlab_blue}{rgb}{0,0.4470,0.7410}
\definecolor{matlab_red}{rgb}{0.8500,0.3250,0.0980}

\begin{document}

\title{Distributionally-Robust Machine Learning \\ Using Locally Differentially-Private Data}

\author{Farhad~Farokhi,~\IEEEmembership{Senior Member,~IEEE}
	\IEEEcompsocitemizethanks{\IEEEcompsocthanksitem F.~Farokhi is with the Department of Electrical and Electronic Engineering at the University of Melbourne, Australia.\protect\\
		E-mail: farhad.farokhi@unimelb.edu.au}
	\thanks{Manuscript received \today.}}


\IEEEtitleabstractindextext{%
	\begin{abstract}
	We consider machine learning, particularly regression, using locally-differentially private datasets. The Wasserstein distance is used to define an ambiguity set centered at the empirical distribution of the dataset corrupted by local differential privacy noise. The ambiguity set is shown to contain the probability distribution of unperturbed, clean data. The radius of the ambiguity set is a function of the privacy budget, spread of the data, and the size of the problem. Hence, machine learning with locally-differentially private datasets can be rewritten as a distributionally-robust optimization. For general distributions, the distributionally-robust optimization problem can relaxed as a regularized machine learning problem with the Lipschitz constant of the machine learning model as a regularizer. For  linear and logistic regression, this regularizer is the dual norm of the model parameters. For Gaussian data, the distributionally-robust optimization problem can be solved exactly to find an optimal regularizer. This approach results in an entirely new regularizer for training linear regression models. Training with this novel regularizer can be posed as a semi-definite program. Finally, the performance of the proposed distributionally-robust machine learning training is demonstrated on practical datasets. 
	\end{abstract}
	
	}

\maketitle

\IEEEdisplaynontitleabstractindextext

\IEEEpeerreviewmaketitle

\IEEEraisesectionheading{\section{Introduction}}
\IEEEPARstart{A}{dvances} in artificial intelligence, particularly  machine learning, have opened new possibilities for data analytic to address important societal challenges. However, these achievements can be stifled by privacy concerns. Local differential privacy, a variant of the popular differential privacy framework~\cite{10100711681878_14,dwork2014algorithmic}, has been touted as an approach for providing privacy guarantees in the presence of an untrusted aggregator or data analysts~\cite{kairouz2016extremal, dewri2013local,6686179}. This is because, with local differential privacy, the data can be freely aggregated and shared. Even, commercial entities, such as Microsoft and Apple, have started using local differential privacy to deploy privacy-preserving data aggregation mechanisms~\cite{ren2018textsf, erlingsson2014rappor, tang2017privacy}.

The additive noise in local differential privacy can degrade the performance of machine learning models trained on perturbed privacy-preserving datasets. Several studies have looked into providing bounds for the performance degradation caused by local differential privacy noise as a function of the privacy budget and the dataset size~\cite{6686179,smith2017interaction, wang2018empirical,zheng2017collect,wang2019noninteractive}. These studies however do not use recent advances in distributionally-robust optimization and machine learning (see, e.g.,~\cite{esfahani2018data,nguyen2018distributionally,kuhn2019wasserstein}) to compute  robust machine learning models with out-of-sample performance guarantees in the presence of local differential privacy noise. They are more concerned by finding the effect of privacy-preserving noise on established machine learning algorithms.

Distributionally-robust optimization considers uncertain stochastic programs~\cite{ben2009robust} with the ambiguity set for the distribution modeled by discrete distributions~\cite{postek2016computationally}, moment constraints~\cite{delage2010distributionally}, Kullback-Leibler divergence~\cite{hu2013kullback}, and the Wasserstein distance~\cite{esfahani2018data}. Distributionally-robust optimization has shown significant promises in adversarial machine learning~\cite{sinha2017certifiable,farokhi2020regularization} or machine learning with outlier data~\cite{chen2018distributionally}. 
However, so far, this approach has not been used for training robust machine learning models based on datasets perturbed using local differential privacy.

In this paper, we use the Wasserstein distance to define an ambiguity set centered at the empirical distribution of the training dataset that is corrupted with local differential privacy noise. This ambiguity set is shown to contain the probability distribution of unperturbed data. The radius of the ambiguity set is a function of the privacy budget, spread of the data, and the size of the problem (i.e., number of inputs and outputs of the machine learning model). Armed with this description of the ambiguity set, we can cast the problem of learning with locally-differentially private data as a distributionally-robust optimization problem. We show that, for general distributions, an upper bound for the worst-case expected loss in the distributionally-robust optimization problem is the empirical sampled-averaged loss plus the Lipschitz-constant of the loss function. Using this, we can relax the distributionally-robust optimization problem as a regularized machine learning problem with the Lipschitz constant as a regularizer. For linear and logistic regression models, this regularizer is equal to the dual norm of the model parameters. For Gaussian data, the distributionally-robust optimization problem can be solved exactly to find an optimal regularizer for the problem. This approach results in an entirely new regularizer for linear regression. We finally demonstrate the performance of the proposed distributionally-robust optimization problems on practical datasets.

\section{Overview of the Wasserstein Distance}\label{sec:Wasserstein}
In this section, a brief overview of the Wasserstein distance is provided. The set of probability distributions $\mathbb{Q}$ over the set $\Xi\subseteq\mathbb{R}^m$ such that $\mathbb{E}^{\mathbb{Q}}\{\|\xi\|\}<\infty$ is denoted by $\mathcal{M}(\Xi)$. For all $\mathbb{P}_1,\mathbb{P}_2\in\mathcal{M}(\Xi)$, the Wasserstein distance $\mathfrak{W}_1:\mathcal{M}(\Xi)\times \mathcal{M}(\Xi)\rightarrow \mathbb{R}_{\geq 0}:=\{x\in \mathbb{R}|x\geq 0\}$ is 
\begin{align*}
\mathfrak{W}_q(\mathbb{P}_1,\mathbb{P}_2):=\inf_{\Pi}& \Bigg\{\left[\int_{\Xi^2} \|\xi_1-\xi_2\|^q\Pi(\mathrm{d}\xi_1,\mathrm{d}\xi_2)\right]^{1/q}: \\
& \Pi \mbox{ is a joint disribution on } \xi_1 \mbox{ and }\xi_2\\
& \mbox{with marginals }\mathbb{P}_1\mbox{ and }\mathbb{P}_2\mbox{, respectively}\Bigg\}.
\end{align*}
The Wasserstein distance is symmetric, i.e.,  $\mathfrak{W}_q(\mathbb{P}_1,\mathbb{P}_2)=\mathfrak{W}_q(\mathbb{P}_2,\mathbb{P}_1)$. The Wasserstein distance satisfies the triangle inequality~\cite[p.\,170]{prugel2020probability}, i.e., $\mathfrak{W}_q(\mathbb{P}_1,\mathbb{P}_3)\leq \mathfrak{W}_q(\mathbb{P}_1,\mathbb{P}_2)+\mathfrak{W}_q(\mathbb{P}_2,\mathbb{P}_3)$. Also, the Wasserstein distance is convex~\cite[Lemma~2.1]{pflug2014multistage}, i.e., $\mathfrak{W}_q(\alpha\mathbb{P}_1+(1-\alpha)\mathbb{P}_2,\mathbb{Q})\leq \alpha \mathfrak{W}_q (\mathbb{P}_1,\mathbb{Q})+(1-\alpha)\mathfrak{W}_q(\mathbb{P}_2,\mathbb{Q})$ for all $\alpha\in[0,1]$. For $q=1$, the duality theorem of Kantorovich and Rubinstein~\cite{kantorovich1958space} implies that 
\begin{align*}
\mathfrak{W}_1(\mathbb{P}_1,\mathbb{P}_2):=\sup_{f\in\mathbb{L}} \Bigg\{&\hspace{-.03in}\int_{\Xi} f(\xi)\mathbb{P}_1(\mathrm{d}\xi)-\int_{\Xi} f(\xi)\mathbb{P}_2(\mathrm{d}\xi)\Bigg\},
\end{align*}
where $\mathbb{L}$ denotes the set of all Lipschitz functions with Lipschitz constant upper bounded by one, i.e., all functions $f$ such that $|f(\xi_1)-f(\xi_2)|\leq \|\xi_1-\xi_2\|$ for all $\xi_1,\xi_2\in\Xi$. 

\section{Distributionally-Robust Machine Learning with Private Data}
\subsection{Expected Risk Minimization}
We consider supervised learning based on a training dataset $\{(x_i,y_i)\}_{i=1}^n$, where $x_i\in\mathcal{X}\subseteq\mathbb{R}^{p_x}$ is the input or feature vector (e.g., pixels of an image or features extracted from it) and $y_i\in\mathcal{Y}\subseteq\mathbb{R}^{p_y}$ is the output or label (e.g., image content). The training dataset is composed of independently and identically distributed (i.i.d.) samples from probability distribution $\mathbb{P}$.

Training machine learning models refers to extracting a model $\mathfrak{M}:\mathbb{R}^{p_x}\times \mathbb{R}^{p_\theta} \rightarrow\mathbb{R}^{p_y}$ (sometimes referred to as  hypothesis) to describe the relationship between inputs and outputs distributed according to $\mathbb{P}$. This can be done by solving the stochastic program
\begin{align} \label{eqn:optimization:1}
J^*:=\min_{\theta\in\Theta}  \mathbb{E}^{\mathbb{P}} \{\ell(\mathfrak{M}(x;\theta),y)\},
\end{align}
where $\theta$ is the machine learning model parameter,  $\Theta\subseteq\mathbb{R}^{p_\theta}$ is the set of feasible parameters, and  $\ell:\mathbb{R}^{p_y}\times \mathbb{R}^{p_y}\rightarrow \mathbb{R}$ is the loss function. An example of a loss function is  $\ell(\mathfrak{M}(x;\theta),y)=(\mathfrak{M}(x;\theta)-y)^\top (\mathfrak{M}(x;\theta)-y)$. The existence of a minimizer in~\eqref{eqn:optimization:1}, and the subsequent approximations in this paper, is guaranteed if the loss function is continuous and the feasible set is compact. This problem is sometimes referred to as expected risk minimization. 

In the absence of the knowledge of the distribution $\mathbb{P}$, the training dataset  $\{(x_i,y_i)\}_{i=1}^n$, i.e., samples from this distributions, can be used to solve the sample-averaged approximation problem
\begin{align} \label{eqn:optimization:2}
\hat{J}:=\min_{\theta\in\Theta}  \frac{1}{n} \sum_{i=1}^n\ell(\mathfrak{M}(x_i;\theta),y_i).
\end{align}
The approximation in~\eqref{eqn:optimization:2} is often the starting point of machine learning. This is because~\eqref{eqn:optimization:2} can be a good proxy for~\eqref{eqn:optimization:1}, when $n$ is large enough, in the sense of probably approximately correct (PAC) learnability~\cite{anthony2009neural}. We make the following standing assumption regarding the distribution of the training data.

\begin{assumption} \label{assum:1} $\mathbb{E}^{\mathbb{P}}\{\exp(\|\xi\|^a)\}<\infty$ for some $a>1$.
\end{assumption}

This assumption implies that $\mathbb{P}$ is a light-tailed distribution. All probability distributions with compact support set are light-tailed; however, unbounded noises, such as Gaussian or Laplace, are also light-tailed. This is often an implicit assumption in the machine learning literature as, for heavy-tailed distributions, the sample average of the loss in~\eqref{eqn:optimization:2} may not even generally converge to the expected loss in~\eqref{eqn:optimization:1} and hence the PAC learnability might not even hold~\cite{brownlees2015empirical, catoni2012challenging}. 

\subsection{Local Differential Privacy}
Due to privacy concerns, the training dataset is sometimes replaced with a noisy dataset $\{(\tilde{x}_i,\tilde{y}_i)\}_{i=1}^n$ in which
\begin{subequations} \label{eqn:LDP}
\begin{align}
\tilde{x}_i:=&x_i+w_i,\label{eqn:LDP:x}\\
\tilde{y}_i:=&y_i,
\end{align}
\end{subequations}
where $(w_i)_{i=1}^n$ are identically and independently distributed (i.i.d.) according to the probability distribution $\mathbb{W}$. Local differential privacy is a useful and versatile notion of privacy. 

\begin{definition}[Local Differential Privacy]
The reporting mechanism with additive noise in~\eqref{eqn:LDP} is $(\epsilon,\delta)$-locally differentially private if, for all $(x,y),(x',y')\in\mathcal{X}\times\mathcal{Y}$ and any Lebesgue-measurable set $\mathcal{A}\subseteq\mathcal{X}\times\mathcal{Y}$,
\begin{align*}
\mathbb{P}\{(\tilde{x}_i,\tilde{y}_i)\in&\mathcal{A}|x_i=x,y_i=y\}\\
&\leq \exp(\epsilon)\mathbb{P}\{(\tilde{x}_i,\tilde{y}_i)\in\mathcal{A}|x_i=x',y_i=y'\}.
\end{align*} 
\end{definition}

\begin{assumption} \label{assume:bound} $\mathcal{X}\subseteq[\underline{x},\overline{x}]^{p_x}$.
\end{assumption}

The box constraint nature of Assumption~\ref{assume:bound} is not strictly-speaking necessary; however, it simplifies the closed-form expression of the results. The results can be readily extended to any compact sets by using diameter of the sets. It is widely known that we can ensure local differential privacy with Laplace and Gaussian additive noises. This is explored in the next theorem. 

\begin{theorem}
	\label{tho:dp} 
	The following statements hold:
	\begin{enumerate}
	\item For $\epsilon>0$, the mechanism in~\eqref{eqn:LDP} is $\epsilon$-locally differentially private if $w_i$ is a vector of zero-mean i.i.d. Laplace noise with scale $\Delta/\epsilon$, where  $\Delta:=(\overline{x}-\underline{x})p_x$;
	\item For $\epsilon,\delta>0$, the mechanism in~\eqref{eqn:LDP} is $(\epsilon,\delta)$-locally differentially private if $w_i$ is a vector of zero-mean i.i.d. Gaussian noise with standard deviation $\sigma:=\sqrt{2\log(1.25/\delta)}\Delta/\epsilon$. 
	\end{enumerate}
\end{theorem}

\begin{IEEEproof}
The proof for the Laplace mechanism follows from~\cite[Theorem~3.6]{dwork2014algorithmic} while noting that the $\ell_1$-sensitivity of the query (which is equal to the identity function for local differential privacy) is given by $\Delta$. The proof for the Gaussian noise follows from~\cite[Theorem~A.1]{dwork2014algorithmic}. Note that the the $\ell_1$-sensitivity is an upper bound for the  $\ell_2$-sensitivity.
\end{IEEEproof}

\subsection{Distributionally-Robust Machine Learning}
The privacy-preserving records in the training dataset $(\tilde{x}_i,\tilde{y}_i)_{i=1}^n$ are independently and identically distributed (i.i.d.) according to $\mathbb{D}$, which can be characterized by the convolution of $\mathbb{P}$ and $\mathbb{W}$. We can define the empirical probability distribution
\begin{align*}
\widehat{\mathbb{D}}_n:=\frac{1}{n}\sum_{i=1}^n \mathfrak{d}_{(\tilde{x}_i,\tilde{y}_i)},
\end{align*}
where $\mathfrak{d}_\xi$ is the Dirac distribution function. Following the definition of the Dirac distribution function, we have
\begin{align}
\frac{1}{n} \sum_{i=1}^n\ell(\mathfrak{M}(x_i;\theta),y_i)=\mathbb{E}^{\widehat{\mathbb{D}}_n} \{\ell(\mathfrak{M}(x;\theta),y)\},
\end{align}
and, as a result, we can rewrite~\eqref{eqn:optimization:2} as 
\begin{align} 
\hat{J}:=\min_{\theta\in\Theta}  \mathbb{E}^{\widehat{\mathbb{D}}_n} \{\ell(\mathfrak{M}(x;\theta),y)\}.
\end{align}
We can show that  the empirical probability distribution $\widehat{\mathbb{D}}_n$ is in a vicinity of the original probability distribution $\mathbb{P}$ with a high probability. 

\begin{theorem} \label{lemma:1} Assume that $\mathbb{W}$ is the distribution in Theorem~\ref{tho:dp}. There exist constants $c_1,c_2>0$ such that 
	\begin{align*}
	\mathbb{D}^n\left\{\mathfrak{W}_1(\widehat{\mathbb{D}}_n,\mathbb{P})
	\leq \zeta(\gamma)
	+\frac{\sqrt{2p}\Delta}{\epsilon}\right\}
	\geq 
	1- \gamma,
	\end{align*}
	for the Laplace mechanism and 
	\begin{align*}
	\mathbb{D}^n\left\{\mathfrak{W}_1(\widehat{\mathbb{D}}_n,\mathbb{P})
	\leq \zeta(\gamma)
	+\frac{\sqrt{2\log(1.25/\delta)p}\Delta}{\epsilon}\right\}
	\geq 
	1- \gamma,
	\end{align*}	
	for the Gaussian mechanism, 
	where
	\begin{align*}
	\zeta(\gamma):=\begin{cases}
	\displaystyle\left(\frac{\log(c_1/\gamma)}{c_2n}\right)^{1/\max\{p,2\}}, & \displaystyle n\geq \frac{\log(c_1/\gamma)}{c_2}, \\
	\displaystyle\left(\frac{\log(c_1/\gamma)}{c_2n}\right)^{1/a}, & \displaystyle n< \frac{\log(c_1/\gamma)}{c_2},
	\end{cases}
	\end{align*}
	for all $n\geq 1$, $p=p_x+p_y\neq 2$, and $\gamma>0$.
\end{theorem}

\begin{IEEEproof}
Note that, since $\mathbb{P}$ is light-tailed, $\mathbb{D}$ is also light-tailed if we use the privacy-preserving noises in Theorem~\ref{tho:dp}. Following~\cite{esfahani2018data}, we know that $\mathbb{D}^n\{\mathfrak{W}_1( \widehat{\mathbb{D}}_n,\mathbb{D}) \leq \zeta(\gamma)\}
	\leq 1-\gamma.$ Using~\cite[Lemma~8.6]{bickel1981some}, we get $\mathfrak{W}_1(\mathbb{D},\mathbb{P})
	\leq \mathfrak{W}_1(\mathbb{P},\mathbb{P}) 
	+\mathfrak{W}_1(\mathfrak{d}_0,\mathbb{W})
	=  \mathfrak{W}_1(\mathfrak{d}_0,\mathbb{W})
	\leq  \mathbb{E}^{\mathbb{W}}\{\|w\|\}
	\leq   \sqrt{\mathbb{E}^{\mathbb{W}}\{\|w\|^2\}},$
	where the last inequality follows from the Jensen's inequality~\cite[p.\,27]{cover2012elements}. Furthermore, for the Laplace noise, we get $\mathbb{E}^{\mathbb{W}}\{\|w\|^2\}=\trace(\mathbb{E}^{\mathbb{W}}\{ww^\top\})=2p\Delta^2/\epsilon^2,$ and, as a result, $\mathfrak{W}_1(\mathbb{D},\mathbb{P}) \leq\sqrt{2p}\Delta/\epsilon.$
	Therefore, $\mathfrak{W}_1(\widehat{\mathbb{D}}_n,\mathbb{P})
	\leq \zeta(\gamma)+\sqrt{2p}\Delta/\epsilon$ if 
	$\mathfrak{W}_1(\widehat{\mathbb{D}}_n,\mathbb{D})
	\leq \zeta(\gamma)$, which implies that $
	\mathbb{D}^n\{\mathfrak{W}_1(\widehat{\mathbb{D}}_n,\mathbb{P})
	\leq \zeta(\gamma)+\sqrt{2p}\mathfrak{d}/\epsilon\}
	\geq 
	1- \gamma.$ The proof for the Gaussian noise is the same with the exception that $\mathbb{E}^{\mathbb{W}}\{\|w\|^2\}=\trace(\mathbb{E}^{\mathbb{W}}\{ww^\top\})={2p\log(1.25/\delta)}\Delta^2/\epsilon^2$.
\end{IEEEproof}

Hence, if we select $\rho$ large enough, the original distribution $\mathbb{P}$ would belong to the ambiguity set $\{\mathbb{G}:\mathfrak{W}_1(\mathbb{G},\widehat{\mathbb{D}}_n)\leq \rho\}$. This observation motivates training the machine learning model by solving the  distributionally-robust optimization problem in
\begin{align} \label{eqn:optimization:3}
\hat{J}_n:=\min_{\theta\in\Theta} \sup_{\mathbb{G}:\mathfrak{W}_1(\mathbb{G},\widehat{\mathbb{D}}_n)\leq \rho}  \mathbb{E}^{\mathbb{G}} \{\ell(\mathfrak{M}(x;\theta),y)\},
\end{align}
for some constant $\rho>0$. The correct value of $\rho$ is discussed in the next theorem.

\begin{theorem} Assume that $\mathbb{W}$ is the distribution in Theorem~\ref{tho:dp}. 
	If  $\rho=\zeta(\beta) +\sqrt{2p}\Delta/\epsilon$ for the Laplace mechanism or if $\rho=\zeta(\beta) +\sqrt{2p\log(1.25/\delta)}\Delta/\epsilon$ for the Gaussian mechanism, then
	\begin{align} \label{eqn:guarantee}
	\mathbb{D}^n\{J^*\leq \mathbb{E}^{\mathbb{P}}\{\ell(\mathfrak{M}(x;\hat{\theta}_n),y)\}\leq \hat{J}_n\}\geq 1-\beta,
	\end{align}	
	where $\beta\in(0,1)$ is a significance parameter and the trained model parameter $\hat{\theta}_n\in\Theta$ is the minimizer of~\eqref{eqn:optimization:3}. 
\end{theorem}

The optimization problem in~\eqref{eqn:optimization:3} involves taking a supremum over the probability density function. This  is an infinite-dimensional optimization problem and is hence computationally difficult to solve. Therefore, we relax this problem in the remainder of this section.

\begin{proposition} \label{lemma:2} Assume that $\ell(\mathfrak{M}(x;\theta),y)$ is $L(\theta)$-Lipschitz continuous in $(x,y)$ for a fixed $\theta\in\Theta$. Then, 
	\begin{align*}
	\sup_{\mathbb{G}:\mathfrak{W}_1(\mathbb{G},\widehat{\mathbb{D}}_n)\leq \rho}  \mathbb{E}^{\mathbb{G}} \{\ell(&\mathfrak{M}(x;\theta),y)\}\\[-1em]
	&\leq \mathbb{E}^{\widehat{\mathbb{D}}_n} \{\ell(\mathfrak{M}(x;\theta),y)\}
	+L(\theta)\rho.
	\end{align*}
\end{proposition}

\begin{IEEEproof}
The proof follows from the duality theorem of Kantorovich and Rubinstein~\cite{kantorovich1958space}.
\end{IEEEproof}

Now, we can define the regularized sample-averaged optimization problem in 
\begin{align} \label{eqn:optimization:4}
\tilde{J}_n:=\min_{\theta\in\Theta}  \Big[\mathbb{E}^{\widehat{\mathbb{D}}_n} \{\ell(\mathfrak{M}(x;\theta),y)\}+\rho L(\theta)\Big].
\end{align}
We can still prove a performance guarantee for the optimizer of~\eqref{eqn:optimization:4}. This is done in the next theorem. 

\begin{theorem} \label{tho:dist_robust_bound} Assume that $\mathbb{W}$ is the distribution in Theorem~\ref{tho:dp}. 
	If  $\rho=\zeta(\beta) +\sqrt{2p}\Delta/\epsilon$ for the Laplace mechanism or if $\rho=\zeta(\beta) +\sqrt{2p\log(1.25/\delta)}\Delta/\epsilon$ for the Gaussian mechanism, then
	\begin{align} 
	\mathbb{D}^n\{J^*\leq \mathbb{E}^{\mathbb{P}}\{\ell(\mathfrak{M}(x;\tilde{\theta}_n),y)\}\leq \tilde{J}_n\}\geq 1-\beta,
	\end{align}	
	where the trained model parameter $\tilde{\theta}_n\in\Theta$ is the minimizer of~\eqref{eqn:optimization:4}. 
\end{theorem}

\begin{IEEEproof}
	The proof is similar to~\cite[Theorem 3.4]{esfahani2018data} with an extra step with the aid of Proposition~\ref{lemma:2}. By selecting $\rho$ as in the statement of the theorem, $\mathbb{P}$ belongs to a ball around $\widehat{\mathbb{D}}_n$ with radius $\rho$ with probability greater than or equal to $1-\beta$ according to Theorem~\ref{lemma:1}. Therefore, with probability of at least $1-\beta$, $\mathbb{E}^{\mathbb{P}} \{\ell(\mathfrak{M}(x;\tilde{\theta}_n),y)\}\leq \sup_{\mathbb{G}:\mathfrak{W}_1(\mathbb{G}, \widehat{\mathbb{D}}_n)\leq \rho} \mathbb{E}^{\mathbb{G}} \{\ell(\mathfrak{M}(x;\tilde{\theta}_n),y)\}$.  Proposition~\ref{lemma:2} states  $\sup_{\mathbb{G}:\mathfrak{W}_1(\mathbb{G}, \widehat{\mathbb{D}}_n)\leq \rho} \mathbb{E}^{\mathbb{G}} \{\ell(\mathfrak{M}(x;\hat{\theta}_n),y)\}\leq \mathbb{E}^{\widehat{\mathbb{D}}_n} \{\ell(\mathfrak{M}(x;\tilde{\theta}),y)\}+\rho L(\tilde{\theta})=\tilde{J}_n$. Therefore, with probability of at least $1-\beta$,  $\mathbb{E}^{\mathbb{P}} \{\ell(\mathfrak{M}(x;\tilde{\theta}_n),y)\}\leq\tilde{J}_n$.\end{IEEEproof}

Theorem~\ref{tho:dist_robust_bound} shows that by regularizing the sample-averaged  cost function, we can train a machine learning model that performs better in the presence of locally differentially private noises. This is an interesting observation demonstrating the value of regularization in privacy machine learning with private data.

\begin{remark}[Large Datasets] In the limit for large $n$, $\zeta(n)\approx 0$. Therefore, following Theorem~\ref{tho:dist_robust_bound},  $\rho=\mathcal{O}(p_x(p_x+p_y)^{1/2}/\epsilon)$ for the Laplace mechanism and $\rho=\mathcal{O}(p_x(p_x+p_y)^{1/2}\log(1/\delta)^{1/2}/\epsilon)$ for the Gaussian mechanism. This implies that the regularization weight $\rho$ should increase when reducing the privacy budget $\epsilon$. Also, higher-dimensional problems, i.e., when $p_x$ or $p_y$ is larger, requires larger regularization weights $\rho$.
\end{remark} 

\begin{remark}[Linear and Logistic Regression] \label{remark:regression}
Without loss of generality, consider $p_y=1$. If  $p_y>1$, each output can be treated independently. In this case, 
$\mathfrak{M}(x;\theta)=\theta^\top 
[
x^\top \;
1
]^\top
$ 
and 
$
\ell(\mathfrak{M}(x;\theta),y)=(\mathfrak{M}(x;\theta)-y)^2/2.$
We also assume that $(x,y)$ belong to compact set $\mathcal{X}\times\mathcal{Y}\subseteq\mathbb{R}^{p_x}\times\mathbb{R}$.
Following~\cite{farokhiDRO2020}, we know that 
$
L(\theta)=(X+1+Y)\|\theta\|_*^2,
$
where  $\|\cdot\|_*$ is the dual norm of $\|\cdot\|$, and
$X=\max_{x\in \mathcal{X}}\|x\|$ and $Y=\max_{y\in\mathcal{Y}}|y|$. An alternative to the quadratic loss function is $\ell(\mathfrak{M}(x;\theta),y)=|\mathfrak{M}(x;\theta)-y|.$
Again, following~\cite{farokhiDRO2020}, 
$L(\theta)=\left\|\theta \right\|_*$.
For the logistic regression, $\mathfrak{M}(x;\theta)=1/({1+\exp(-[x^\top \; 1]\theta)})$ and 
$\ell(\mathfrak{M}(x;\theta),y)=-y\log(\mathfrak{M}(x;\theta))
-(1-y)\log(1-\mathfrak{M}(x;\theta)).
$ Following~\cite{farokhiDRO2020}, $
L(\theta)=(Y+X+2)\|\theta\|_*.$
\end{remark}

\section{Linear Regression with Gaussian Data} \label{sec:optimal_regression}
In this section, we consider the specific case that $(x_i,y_i)_{i=1}^n$ is  Gaussian distributed with mean $\mu$ and covariance $\Sigma$. Furthermore, we assume that $\mathfrak{M}(x;\theta)=Ax+B$ with $A,B$ modeling the machine learning model parameters (instead of $\theta$) and   $\ell(\mathfrak{M}(x;A,B),y)=\|\mathfrak{M}(x;A,B)-y\|^2$. Therefore, by using the Gaussian mechanism in Theorem~\ref{tho:dp}, $(\tilde{x}_i,\tilde{y}_i)_{i=1}^n$ is also Gaussian distributed. Note that Assumption~\ref{assume:bound} no longer holds in this section (as the Gaussian process behind the data has an infinite support). However, with high probability, the data belongs to a bounded set. Therefore, we can adopt local randomized differential privacy instead of local differential privacy~\cite{machanavajjhala2008privacy, rubinstein2017pain,hall2012random}.

In the Gaussian linear regression case described above, we can redefine define the empirical probability distribution $\widehat{\mathbb{D}}_n$ to be Gaussian with mean $\widehat{\mu}_n$ and covariance $\widehat{\Sigma}_n$, where
\begin{align*}
\widehat{\mu}_n:=&\frac{1}{n} \sum_{i=1}^n \begin{bmatrix}
\tilde{x}_i \\ \tilde{y}_i
\end{bmatrix},\\
\widehat{\Sigma}_n:=&\frac{1}{n-1} \sum_{i=1}^n \left(\begin{bmatrix}
\tilde{x}_i \\ \tilde{y}_i
\end{bmatrix}-\widehat{\mu}_n\right)\left(\begin{bmatrix}
\tilde{x}_i \\ \tilde{y}_i
\end{bmatrix}-\widehat{\mu}_n\right)^\top.
\end{align*}

\begin{theorem} \label{lemma:1_1} Assume that $\mathbb{W}$ is the Gaussian distribution in Theorem~\ref{tho:dp}. With high probability, for any $\varepsilon>0$, there exists $n_\varepsilon\geq 0$ such that if $n\geq n_\varepsilon$,
	\begin{align*}
	\mathfrak{W}_2(\widehat{\mathbb{D}}_n,\mathbb{P})
	\leq \varepsilon+ \frac{\sqrt{2\log(1.25/\delta)p}\Delta}{\epsilon}.
	\end{align*}
\end{theorem}

\begin{IEEEproof}
The proof is similar to Theorem~\ref{lemma:1} with the exception of using the fact that $\mathfrak{W}_2 ( \widehat{\mathbb{D}}_n,\mathbb{D})\rightarrow 0$ with probability one as $n\rightarrow\infty$~\cite[Theorem 2.1]{rippl2016limit}. 
\end{IEEEproof}

In this section, we consider the big data regime ($n\gg 1$) so, without loss of generality, $\varepsilon\approxeq 0$ and $\mu\approxeq \widehat{\mu}_n$. Therefore, by using Theorem~\ref{lemma:1_1}, the original distribution $\mathbb{P}$ would belong to the ambiguity set $\{\mathbb{G}:\mathbb{G} \mbox{ is Gaussian}\wedge \mathbb{E}^{\mathbb{G}}\{(x,y)\}=\widehat{\mu}_n \wedge\mathfrak{W}_2(\mathbb{G},\widehat{\mathbb{D}}_n)\leq \rho\}$  if we select $\rho=\sqrt{2\log(1.25/\delta)p}\Delta/\epsilon$. This observation motivates training the machine learning model by solving the  distributionally-robust optimization problem in
\begin{align} \label{eqn:optimization:5}
\hat{J}_n:=\min_{A,B} \sup_{\scriptsize
\begin{array}{c}
\scriptsize
\mathbb{G}:
\scriptsize \mathbb{G}\mbox{ is Gaussian} \\
\mathbb{E}^{\mathbb{G}}\{(x,y)\}=\widehat{\mu}_n \\
\mathfrak{W}_2(\mathbb{G},\widehat{\mathbb{D}}_n)\leq \rho
\end{array}
}  \mathbb{E}^{\mathbb{G}} \{\ell(\mathfrak{M}(x;A,B),y)\},
\end{align}
for constant $\rho=\sqrt{2\log(1.25/\delta)p}\Delta/\epsilon$. 
Following~\cite[Proposition~7]{givens1984class}, if $\widehat{\mathbb{D}}_n$ and $\mathbb{G}$ are both Gaussian with same mean $\mu\approxeq \widehat{\mu}_n$, we get
\begin{align} 
\mathfrak{W}_2 ( \widehat{\mathbb{D}}_n,\mathbb{G})^2
=&\trace(\Sigma_G+\widehat{\Sigma}_n-2(\widehat{\Sigma}_n^{1/2}{\Sigma_G}\widehat{\Sigma}_n^{1/2})^{1/2}),
\label{eqn:GaussianWasserstein}
\end{align}
where $\Sigma_G$ denotes the covariance of $\mathbb{G}$. The expected risk is given by
\begin{align*}
\mathbb{E}^{\mathbb{G}} &\{\ell(\mathfrak{M}(x;A,B),y)\}\\
&=\trace(\mathbb{E}^{\mathbb{G}} \{(Ax+B-y)(Ax+B-y)^\top \})\\
&=\trace\Bigg(\mathbb{E}^{\mathbb{G}} \Bigg\{
\begin{bmatrix}
A & -I 
\end{bmatrix}
\begin{bmatrix}
x \\ y
\end{bmatrix}
\begin{bmatrix}
x \\ y
\end{bmatrix}^\top 
\begin{bmatrix}
A & -I 
\end{bmatrix}^\top 
+BB^\top\\
&\hspace{.5in}+\begin{bmatrix}
A & -I 
\end{bmatrix} 
\begin{bmatrix}
x \\ y
\end{bmatrix}
B^\top
+
B
\begin{bmatrix}
x \\ y
\end{bmatrix}^\top
\begin{bmatrix}
A & -I 
\end{bmatrix}^\top \Bigg\}\Bigg)\\
&=
\trace\Big( 
\begin{bmatrix}
A & -I 
\end{bmatrix}
(\Sigma_G+\widehat{\mu}_n\widehat{\mu}_n^\top )
\begin{bmatrix}
A & -I 
\end{bmatrix}^\top 
+BB^\top\\
&\hspace{.5in}+\begin{bmatrix}
A & -I 
\end{bmatrix} 
\widehat{\mu}_n
B^\top
+
B
\widehat{\mu}_n^\top
\begin{bmatrix}
A & -I 
\end{bmatrix}^\top \Big).
\end{align*}
Using~\cite[Proposition~2.8]{nguyen2018distributionally}, it can be deduced that
\begin{align*}
\hspace*{.8in}&\hspace*{-.8in}\sup_{\scriptsize
	\begin{array}{c}
	\scriptsize
	\mathbb{G}:
	\scriptsize \mathbb{G}\mbox{ is Gaussian} \\
	\mathbb{E}^{\mathbb{G}}\{(x,y)\}=\widehat{\mu}_n \\
	\mathfrak{W}_2(\mathbb{G},\widehat{\mathbb{D}}_n)\leq \rho
	\end{array}
}  \mathbb{E}^{\mathbb{G}} \{\ell(\mathfrak{M}(x;A,B),y)\}
\\
=&\trace\Big( \begin{bmatrix}
A & -I 
\end{bmatrix}
\widehat{\mu}_n\widehat{\mu}_n^\top
\begin{bmatrix}
A & -I 
\end{bmatrix}^\top \\
&\hspace{.5in}+BB^\top+\begin{bmatrix}
A & -I 
\end{bmatrix} 
\widehat{\mu}_n
B^\top
\\&\hspace{.5in}+
B
\widehat{\mu}_n^\top
\begin{bmatrix}
A & -I 
\end{bmatrix}^\top\Big)
+f(A),
\end{align*}
where 
\begin{align*}
f(A):=\inf_{\xi }&\;\Bigg[\xi (\rho^2-\trace(\widehat{\Sigma}_n))\\
\hspace{.8in}&+\xi^2\trace\Bigg(\Bigg(\xi I- \begin{bmatrix}
A^\top \\ -I 
\end{bmatrix} \begin{bmatrix}
A & -I 
\end{bmatrix}\Bigg)^{-1}\widehat{\Sigma}_n\Bigg)\Bigg],\\
\mathrm{s.t.}&\;\xi I\succ \begin{bmatrix}
A^\top \\ -I 
\end{bmatrix} \begin{bmatrix}
A & -I 
\end{bmatrix} .
\end{align*}
Further,
\begin{align*}
 \trace\Big(&  \begin{bmatrix}
 A & -I 
 \end{bmatrix}
 \widehat{\mu}_n\widehat{\mu}_n^\top
 \begin{bmatrix}
 A & -I 
 \end{bmatrix}^\top +BB^\top\\
 &+\begin{bmatrix}
 A & -I 
 \end{bmatrix} 
 \widehat{\mu}_n
 B^\top+
 B
 \widehat{\mu}_n^\top
 \begin{bmatrix}
 A & -I 
 \end{bmatrix}^\top\Big)\\
 &\hspace{-.2in}=\mathbb{E}^{\widehat{\mathbb{D}}_n} \{\ell(\mathfrak{M}(x;A,B),y)\}-\begin{bmatrix}
 A & -I 
 \end{bmatrix}
 \widehat{\Sigma}_n
 \begin{bmatrix}
 A & -I 
 \end{bmatrix}^\top .
\end{align*}
Therefore, the optimization problem in~\eqref{eqn:optimization:5} can be rewritten as 
\begin{align} \label{eqn:optimization:6}
\hat{J}_n:=\min_{A,B} \mathbb{E}^{\widehat{\mathbb{D}}_n} \{\ell(\mathfrak{M}(x;A,B),y)\}+\lambda(A),
\end{align}
where $
\lambda(A):=f(A)-\begin{bmatrix}
A & -I 
\end{bmatrix}
\widehat{\Sigma}_n
\begin{bmatrix}
A & -I 
\end{bmatrix}^\top $
is the optimal regularization for linear regression with locally-differential private data with Gaussian data. This regularization is completely novel in the context of linear regression. In what follows, we provide a more tractable formulation for~\eqref{eqn:optimization:6}. 

\begin{theorem}
The solution to~\eqref{eqn:optimization:6} is given by
\begin{subequations}
\begin{align}
\min_{A,B,\xi,Z}\;&\xi (\rho^2-\trace(\widehat{\Sigma}_n))+\trace(Z)\nonumber\\
&+\trace\Big( \begin{bmatrix}
A & -I 
\end{bmatrix}
\widehat{\mu}_n\widehat{\mu}_n^\top
\begin{bmatrix}
A & -I 
\end{bmatrix}^\top + BB^\top\nonumber\\
&\hspace{.2in}+\begin{bmatrix}
A & -I 
\end{bmatrix} 
\widehat{\mu}_n
B^\top+
B
\widehat{\mu}_n^\top
\begin{bmatrix}
A & -I 
\end{bmatrix}^\top\Big),\\
\mathrm{s.t.} \quad& 
\begin{bmatrix}
Z & \widehat{\Sigma}_n^{1/2}\xi & 0 \\
\widehat{\Sigma}_n^{1/2}\xi & \xi I & 
\begin{bmatrix}
A^\top \\ -I 
\end{bmatrix} \\
0 & \begin{bmatrix}
A & -I 
\end{bmatrix} & I
\end{bmatrix}\succeq 0.
\end{align}
\end{subequations}
\end{theorem}

\begin{IEEEproof}
Let $Z\succeq 0$ be such that 
\begin{align*}
\xi^2\widehat{\Sigma}_n^{1/2}\Bigg(\xi I- \begin{bmatrix}
A^\top \\ -I 
\end{bmatrix} \begin{bmatrix}
A & -I 
\end{bmatrix}\Bigg)^{-1}\widehat{\Sigma}_n^{1/2}\preceq Z.
\end{align*}
Using the Schur complement~\cite{zhang2006schur}, we can transform this inequality into 
\begin{align*}
\begin{bmatrix}
Z & \widehat{\Sigma}_n^{1/2}\xi \\
\widehat{\Sigma}_n^{1/2}\xi & \xi I- \begin{bmatrix}
A^\top \\ -I 
\end{bmatrix} \begin{bmatrix}
A & -I 
\end{bmatrix}
\end{bmatrix}\succeq 0.
\end{align*}
Again, using the Schur complement, this matrix inequality can be transformed into
\begin{align*}
\begin{bmatrix}
Z & \widehat{\Sigma}_n^{1/2}\xi \\
\widehat{\Sigma}_n^{1/2}\xi & \xi I
\end{bmatrix}
- \begin{bmatrix}
0 \\ A^\top \\ -I 
\end{bmatrix} \begin{bmatrix}
0 & A & -I 
\end{bmatrix}
\end{align*}
\begin{align}\label{eqn:LMI1}
\begin{bmatrix}
Z & \widehat{\Sigma}_n^{1/2}\xi & 0 \\
\widehat{\Sigma}_n^{1/2}\xi & \xi I & 
\begin{bmatrix}
A^\top \\ -I 
\end{bmatrix} \\
0 & \begin{bmatrix}
A & -I 
\end{bmatrix} & I
\end{bmatrix}\succeq 0.
\end{align}
Finally, using the Schur complement, the constraint in computing $f(A)$ can rewritten as
\begin{align} \label{eqn:LMI2}
\begin{bmatrix}
\xi I &  \begin{bmatrix}
A^\top \\ -I
\end{bmatrix} \\
\begin{bmatrix}
A & -I
\end{bmatrix}
 & I
\end{bmatrix}\succeq 0.
\end{align}
Note that~\eqref{eqn:LMI2} is a subset of~\eqref{eqn:LMI1} and thus need not be added to the constraints. 
\end{IEEEproof}

\begin{figure}
	\centering
	\begin{tikzpicture}
	\node[] at (0,0) {\includegraphics[width=0.9\linewidth]{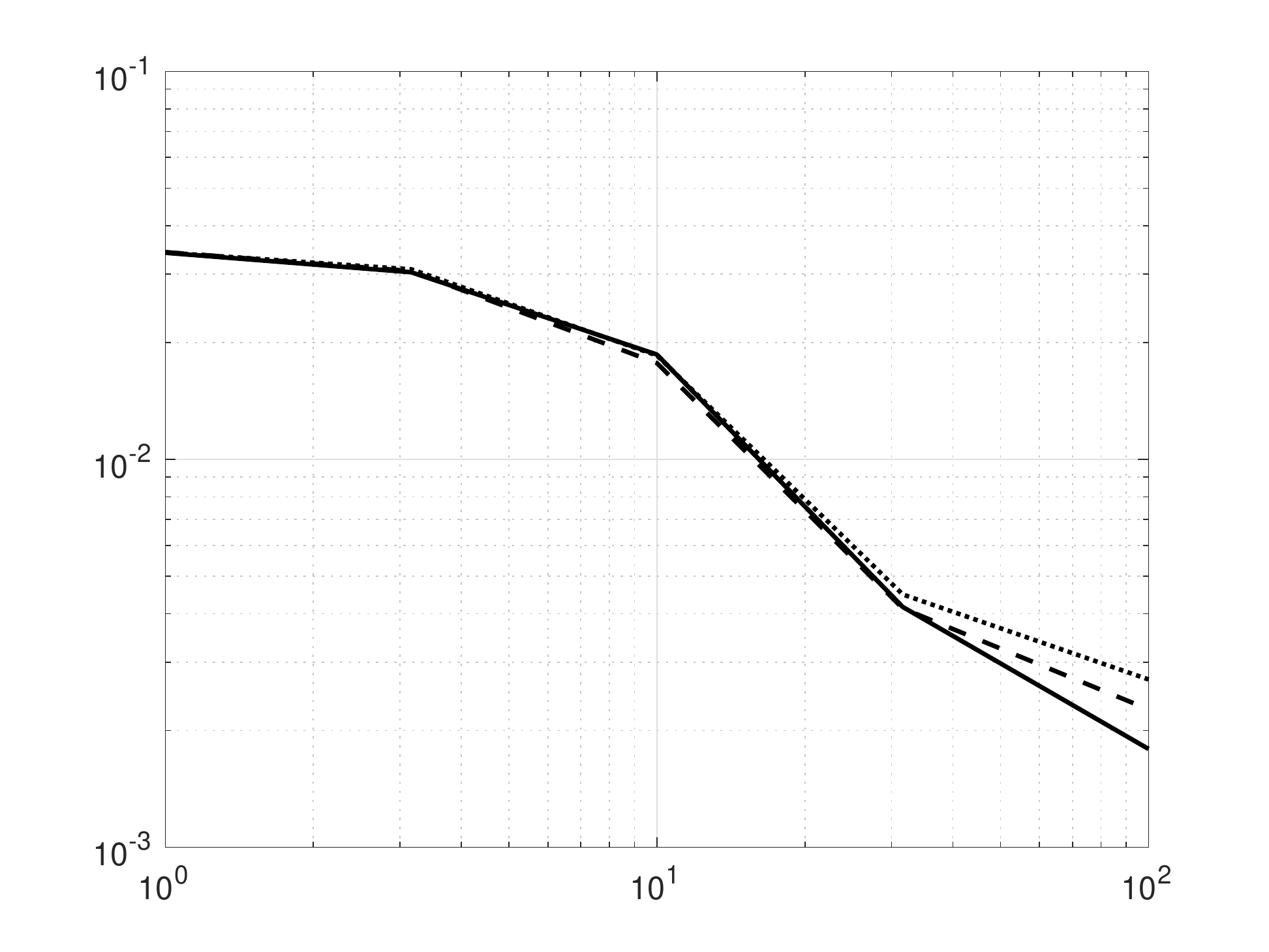}};
	\node[] at (0,-3.0) {$\epsilon$};
	\node[rotate=90] at (-3.9,0) {$ \mathbb{E}^{\mathbb{P}} \{\ell(\mathfrak{M}(x;A,B),y)\}$};
	\node[draw,minimum width=2.9cm,minimum height=1.1cm,fill=white] at (1.65,1.9) {}; 
	\draw[-,line width=1] (.4,2.22) -- (.7,2.22);
	\draw[-,dashed,line width=1] (.4,1.9) -- (.7,1.9);
	\draw[-,dotted,line width=1.5] (.4,1.58) -- (.7,1.58);
	\node[] at (2.3,1.9) {
		\begin{minipage}{3cm} \footnotesize
		\eqref{eqn:optimization:6}\\
		\eqref{eqn:optimization:4} with $\rho=10^{-2}$ \\
		\eqref{eqn:optimization:2}
		\end{minipage}
	};
	\end{tikzpicture}
	\vspace*{-.2in}
	\caption{
		\label{fig:1} Performance of the linear regression for the loan dataset trained on the locally-differential private dataset tested on the original probability distribution.
	}
\end{figure}

\begin{figure}
	\centering
	\begin{tikzpicture}
	\node[] at (0,0) {\includegraphics[width=.9\linewidth]{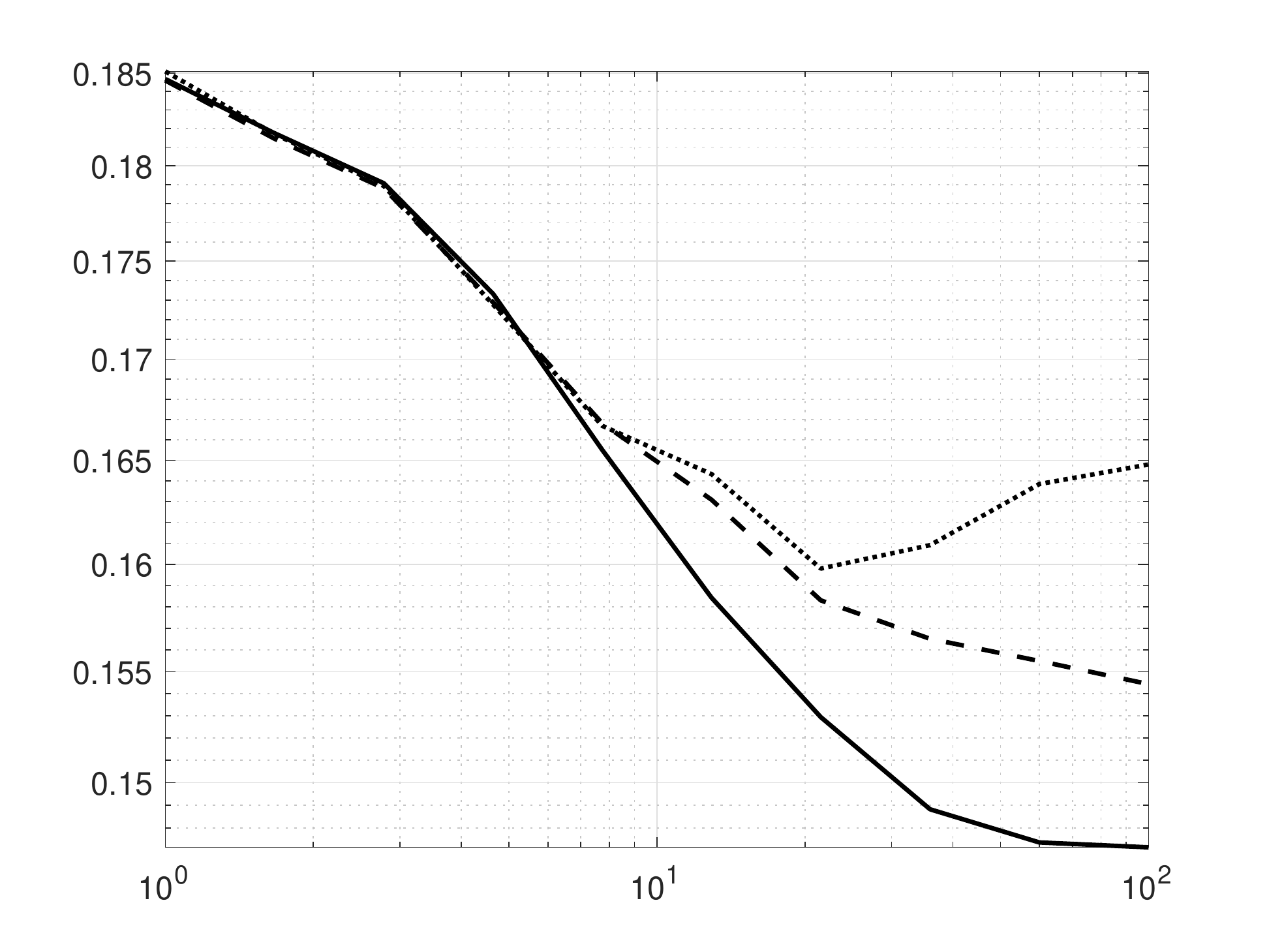}};
	\node[] at (0,-3.0) {$\epsilon$};
	\node[rotate=90] at (-3.9,0) {$ \mathbb{E}^{\mathbb{P}} \{\ell(\mathfrak{M}(x;A,B),y)\}$};
	\node[draw,minimum width=2.9cm,minimum height=1.1cm,fill=white] at (1.65,1.9) {}; 
	\draw[-,line width=1] (.4,2.22) -- (.7,2.22);
	\draw[-,dashed,line width=1] (.4,1.9) -- (.7,1.9);
	\draw[-,dotted,line width=1.5] (.4,1.58) -- (.7,1.58);
	\node[] at (2.3,1.9) {
		\begin{minipage}{3cm} \footnotesize
		\eqref{eqn:optimization:6}\\
		\eqref{eqn:optimization:4} with $\rho=10^{-2}$ \\
		\eqref{eqn:optimization:2}
		\end{minipage}
	};
	\end{tikzpicture}
	\vspace*{-.2in}
	\caption{
		\label{fig:2} Performance of the linear regression for the Adult dataset trained on the locally-differential private dataset tested on the original probability distribution.
	}
\end{figure}
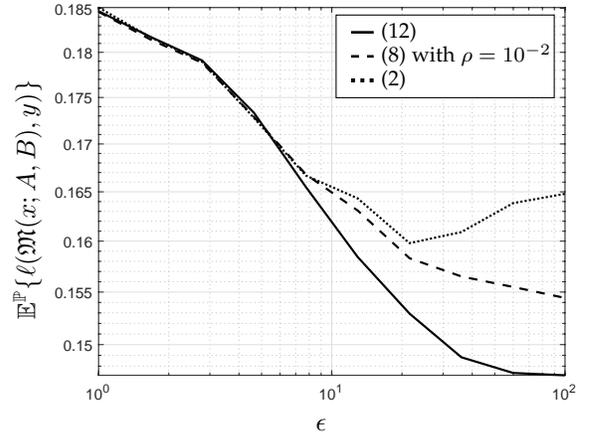

\section{Experimental Result}
Here, we demonstrate the performance of  distributionally-robust machine learning on two practical datasets. 
\subsection{Loan Dataset} The dataset contains information of roughly 887,000 loans~\cite{kaggle1}. The inputs contain loan information, e.g., total loan size, and borrower information, e.g.,  age. The outputs are the interest rates of loans. Categorical features, e.g., state of residence and loan grade, are encoded with integer numbers. Unique identifiers, e.g., identity, and irrelevant attributes, e.g., URLs, are removed. We scale all input attributes and outputs to be between zero and one to meet Assumption~\ref{assume:bound}. We consider linear regression framework in Remark~\ref{remark:regression} and Section~\ref{sec:optimal_regression}. We use the Gaussian mechanism in Theorem~\ref{tho:dp} to generate locally-differentially private datasets with $\delta=10^{-2}$ and $\epsilon\in[10^{0},10^2]/p_x$. We use 50 entries of the dataset for training and the remaining entries for evaluation. Note that we are using such a low number of data entries as we are using linear regression. Figure~\ref{fig:1} illustrates the performance of the linear regression for the loan dataset trained on the locally-differential private dataset tested on the original probability distribution. Regularization clearly improves the out-of-sample performance of the model. Furthermore, the optimal regularization in Section~\ref{sec:optimal_regression} can be better than the generic regularization based on the Lipschitz constant of the loss function in~\eqref{eqn:optimization:4}. 

\subsection{Adult Dataset}
This dataset contains nearly 49,000 records from the 1994 Census database~\cite{Dua:2017}. The records contain features, e.g., age and education. The output is binary number indicating whether an individuals earns more than \$50,000. Similarly, we scale all input attributes and outputs to be between zero and one in line with Assumption~\ref{assume:bound}. We consider the linear regression framework in Remark~\ref{remark:regression} and Section~\ref{sec:optimal_regression}. We use the Gaussian mechanism in Theorem~\ref{tho:dp} to generate locally-differentially private datasets with $\delta=10^{-2}$ and $\epsilon\in[10^{0},10^2]/p_x$. We use 50 entries of the dataset for training and the remaining entries for evaluation. Again, we are using such a low number of data entries as we are using linear regression. Figure~\ref{fig:2} illustrates the performance of the linear regression for the adult dataset trained on the locally-differential private dataset tested on the original probability distribution. Regularization clearly improves the performance of the model.

\section{Conclusions}
We considered machine learning, particularly regression, using locally-differentially private datasets is considered. We posed machine learning with locally-differentially private datasets as a distributionally-robust optimization with an ambiguity set parameterized by the Wasserstein distance. For general distributions, the distributionally-robust optimization problem was relaxed as a regularized machine learning problem with the Lipschitz constant of the machine learning model as a regularizer. For Gaussian data, the distributionally-robust optimization problem was solved exactly to find an optimal regularizer. 

\bibliographystyle{ieeetr}
\bibliography{citation}

\begin{thebibliography}{10}

\bibitem{10100711681878_14}
C.~Dwork, F.~McSherry, K.~Nissim, and A.~Smith, ``Calibrating noise to
  sensitivity in private data analysis,'' in {\em Theory of Cryptography}
  (S.~Halevi and T.~Rabin, eds.), (Berlin, Heidelberg), pp.~265--284, Springer
  Berlin Heidelberg, 2006.

\bibitem{dwork2014algorithmic}
C.~Dwork and A.~Roth, ``The algorithmic foundations of differential privacy,''
  {\em Foundations and Trends{\textregistered} in Theoretical Computer
  Science}, vol.~9, no.~3--4, pp.~211--407, 2014.

\bibitem{kairouz2016extremal}
P.~Kairouz, S.~Oh, and P.~Viswanath, ``Extremal mechanisms for local
  differential privacy,'' {\em The Journal of Machine Learning Research},
  vol.~17, no.~1, pp.~492--542, 2016.

\bibitem{dewri2013local}
R.~Dewri, ``Local differential perturbations: Location privacy under
  approximate knowledge attackers,'' {\em IEEE Transactions on Mobile
  Computing}, vol.~12, no.~12, pp.~2360--2372, 2013.

\bibitem{6686179}
J.~C. Duchi, M.~I. Jordan, and M.~J. Wainwright, ``Local privacy and
  statistical minimax rates,'' in {\em 2013 IEEE 54th Annual Symposium on
  Foundations of Computer Science}, pp.~429--438, 2013.

\bibitem{ren2018textsf}
X.~Ren, C.-M. Yu, W.~Yu, S.~Yang, X.~Yang, J.~A. McCann, and S.~Y. Philip,
  ``{LoPub}: High-dimensional crowdsourced data publication with local
  differential privacy,'' {\em IEEE Transactions on Information Forensics and
  Security}, vol.~13, no.~9, pp.~2151--2166, 2018.

\bibitem{erlingsson2014rappor}
{\'U}.~Erlingsson, V.~Pihur, and A.~Korolova, ``{RAPPOR}: Randomized
  aggregatable privacy-preserving ordinal response,'' in {\em Proceedings of
  the 2014 ACM SIGSAC conference on computer and communications security},
  pp.~1054--1067, 2014.

\bibitem{tang2017privacy}
J.~Tang, A.~Korolova, X.~Bai, X.~Wang, and X.~Wang, ``Privacy loss in {Apple}'s
  implementation of differential privacy on {MacOS 10.12},'' {\em arXiv
  preprint arXiv:1709.02753}, 2017.

\bibitem{smith2017interaction}
A.~Smith, A.~Thakurta, and J.~Upadhyay, ``Is interaction necessary for
  distributed private learning?,'' in {\em 2017 IEEE Symposium on Security and
  Privacy (SP)}, pp.~58--77, IEEE, 2017.

\bibitem{wang2018empirical}
D.~Wang, M.~Gaboardi, and J.~Xu, ``Empirical risk minimization in
  non-interactive local differential privacy revisited,'' in {\em Advances in
  Neural Information Processing Systems}, pp.~965--974, 2018.

\bibitem{zheng2017collect}
K.~Zheng, W.~Mou, and L.~Wang, ``Collect at once, use effectively: making
  non-interactive locally private learning possible,'' in {\em Proceedings of
  the 34th International Conference on Machine Learning-Volume 70},
  pp.~4130--4139, 2017.

\bibitem{wang2019noninteractive}
D.~Wang, A.~Smith, and J.~Xu, ``Noninteractive locally private learning of
  linear models via polynomial approximations,'' in {\em Algorithmic Learning
  Theory}, pp.~898--903, 2019.

\bibitem{esfahani2018data}
P.~M. Esfahani and D.~Kuhn, ``Data-driven distributionally robust optimization
  using the wasserstein metric: Performance guarantees and tractable
  reformulations,'' {\em Mathematical Programming}, vol.~171, no.~1-2,
  pp.~115--166, 2018.

\bibitem{nguyen2018distributionally}
V.~A. Nguyen, D.~Kuhn, and P.~M. Esfahani, ``Distributionally robust inverse
  covariance estimation: The wasserstein shrinkage estimator,'' {\em arXiv
  preprint arXiv:1805.07194}, 2018.

\bibitem{kuhn2019wasserstein}
D.~Kuhn, P.~M. Esfahani, V.~A. Nguyen, and S.~Shafieezadeh-Abadeh,
  ``Wasserstein distributionally robust optimization: Theory and applications
  in machine learning,'' in {\em Operations Research \& Management Science in
  the Age of Analytics}, pp.~130--166, INFORMS, 2019.

\bibitem{ben2009robust}
A.~Ben-Tal, L.~El~Ghaoui, and A.~Nemirovski, {\em Robust optimization},
  vol.~28.
\newblock Princeton University Press, 2009.

\bibitem{postek2016computationally}
K.~Postek, D.~den Hertog, and B.~Melenberg, ``Computationally tractable
  counterparts of distributionally robust constraints on risk measures,'' {\em
  SIAM Review}, vol.~58, no.~4, pp.~603--650, 2016.

\bibitem{delage2010distributionally}
E.~Delage and Y.~Ye, ``Distributionally robust optimization under moment
  uncertainty with application to data-driven problems,'' {\em Operations
  research}, vol.~58, no.~3, pp.~595--612, 2010.

\bibitem{hu2013kullback}
Z.~Hu and L.~J. Hong, ``Kullback-leibler divergence constrained
  distributionally robust optimization,'' {\em Available at Optimization
  Online}, 2013.

\bibitem{sinha2017certifiable}
A.~Sinha, H.~Namkoong, and J.~Duchi, ``Certifiable distributional robustness
  with principled adversarial training,'' in {\em Proceedings of the Machine
  Learning and Computer Security Workshop (co-located with Conference on Neural
  Information Processing Systems 2017)}, vol.~2, 2017.

\bibitem{farokhi2020regularization}
F.~Farokhi, ``Regularization helps with mitigating poisoning attacks:
  Distributionally-robust machine learning using the wasserstein distance,''
  {\em arXiv preprint arXiv:2001.10655}, 2020.

\bibitem{chen2018distributionally}
R.~Chen and I.~C. Paschalidis, ``A distributionally robust optimization
  approach for outlier detection,'' in {\em 2018 IEEE Conference on Decision
  and Control (CDC)}, pp.~352--357, IEEE, 2018.

\bibitem{prugel2020probability}
A.~Pr{\"u}gel-Bennett, {\em The Probability Companion for Engineering and
  Computer Science}.
\newblock Cambridge University Press, 2020.

\bibitem{pflug2014multistage}
G.~C. Pflug and A.~Pichler, {\em Multistage Stochastic Optimization}.
\newblock Springer Series in Operations Research and Financial Engineering,
  Springer International Publishing, 2014.

\bibitem{kantorovich1958space}
L.~V. Kantorovich and G.~Rubinshtein, ``On a space of totally additive
  functions,'' {\em Vestn. Lening. Univ}, vol.~13, pp.~52--59, 1958.

\bibitem{anthony2009neural}
M.~Anthony and P.~L. Bartlett, {\em Neural Network Learning: Theoretical
  Foundations}.
\newblock Cambridge University Press, 2009.

\bibitem{brownlees2015empirical}
C.~Brownlees, E.~Joly, G.~Lugosi, {\em et~al.}, ``Empirical risk minimization
  for heavy-tailed losses,'' {\em The Annals of Statistics}, vol.~43, no.~6,
  pp.~2507--2536, 2015.

\bibitem{catoni2012challenging}
O.~Catoni, ``Challenging the empirical mean and empirical variance: A deviation
  study,'' in {\em Annales de l'Institut Henri Poincar\'{e}, Probabilit{\'e}s
  et Statistiques}, vol.~48, pp.~1148--1185, 2012.

\bibitem{bickel1981some}
P.~J. Bickel, D.~A. Freedman, {\em et~al.}, ``Some asymptotic theory for the
  bootstrap,'' {\em The annals of statistics}, vol.~9, no.~6, pp.~1196--1217,
  1981.

\bibitem{cover2012elements}
T.~M. Cover and J.~A. Thomas, {\em Elements of Information Theory}.
\newblock Wiley, 2012.

\bibitem{farokhiDRO2020}
F.~Farokhi, ``Regularization helps with mitigating poisoning attacks:
  Distributionally-robust machine learning using the wasserstein distance,''
  2020.
\newblock arXiv:2001.10655, \url{https://arxiv.org/abs/2001.10655}.

\bibitem{machanavajjhala2008privacy}
A.~Machanavajjhala, D.~Kifer, J.~Abowd, J.~Gehrke, and L.~Vilhuber, ``Privacy:
  Theory meets practice on the map,'' in {\em Proceedings of the 2008 IEEE 24th
  International Conference on Data Engineering}, pp.~277--286, 2008.

\bibitem{rubinstein2017pain}
B.~I. Rubinstein and F.~Ald{\`a}, ``Pain-free random differential privacy with
  sensitivity sampling,'' in {\em International Conference on Machine
  Learning}, pp.~2950--2959, 2017.

\bibitem{hall2012random}
R.~Hall, A.~Rinaldo, and L.~Wasserman, ``Random differential privacy,'' {\em
  Journal of Privacy and Confidentiality}, vol.~4, no.~2, pp.~43--59, 2012.

\bibitem{rippl2016limit}
T.~Rippl, A.~Munk, and A.~Sturm, ``Limit laws of the empirical wasserstein
  distance: Gaussian distributions,'' {\em Journal of Multivariate Analysis},
  vol.~151, pp.~90--109, 2016.

\bibitem{givens1984class}
C.~R. Givens and R.~M. Shortt, ``A class of wasserstein metrics for probability
  distributions.,'' {\em The Michigan Mathematical Journal}, vol.~31, no.~2,
  pp.~231--240, 1984.

\bibitem{zhang2006schur}
F.~Zhang, {\em The Schur complement and its applications}, vol.~4.
\newblock Springer Science \& Business Media, 2006.

\bibitem{kaggle1}
W.~Kan, ``Lending club loan data: Analyze lending club's issued loans,'' 2016.
\newblock Kaggle, \url{https://www.kaggle.com/wendykan/lending-club-loan-data}.

\bibitem{Dua:2017}
D.~Dheeru and E.~Karra~Taniskidou, ``{UCI} machine learning repository,'' 2017.
\newblock University of California, Irvine,
  \url{http://archive.ics.uci.edu/ml}.

\end{thebibliography}

\end{document}